# Improving robot understanding using conversational AI: demonstration and feasibility study*

Shikhar Kumar and Yael Edan

*Abstract*— Explanations constitute an important aspect of successful human–robot interactions and can enhance robot understanding. To improve the understanding of the robot, we have developed four levels of explanation (LOE) based on two questions: *what* needs to be explained, and *why* the robot has made a particular decision. The understandable robot requires a communicative action when there is disparity between the human's mental model of the robot and the robot's state of mind. This communicative action was generated by utilizing a conversational AI platform to generate explanations. An adaptive dialog was implemented for transition from one LOE to another. Here, we demonstrate the adaptive dialog in a collaborative task with errors and provide results of a feasibility study with users.

## I. INTRODUCTION

For the widespread deployment of autonomous robots in everyday tasks, the actions, decisions and intentions of the robots must be understood by the users. Understandability in human-robot interaction (HRI) may be defined as the extent to which humans can understand the robot and its actions [1]. The importance of understandability lies in its ability to improve the efficiency and safety of collaborative tasks and to positively impact HRI [2]–[4] and task performance [5].

The model of the 'understandable' robot [1] is based on the fact that each agent in HRI (i.e., the robot and the human) has both its own state of mind and a mental model of the interacting agent. A communicative action needs to be generated when there is a disparity between the robot's state of mind and the mental model of the robot in the human's state of mind.

The above communicative action is the verbal explanation that is the focus of this work. To lessen the disparity between the state of mind of the robot and the human's mental model of the robot, we designed this verbal communication to be an adaptive system. The verbalization of explanations was implemented by developing levels of explanation (LOEs) [6]. The underlying concept is that the LOEs will enhance communication efficiency by providing a brief explanation if the context of the task is known; otherwise, a detailed explanation will be provided. This paper demonstrates the development of the LOEs along with their implementation in an adaptive dialog system in a collaborative human-robot task.

## II. DESIGN OF LEVELS OF EXPLANATION

The development of the LOEs was based on two questions, namely: (1) *what* the robot needs to explain, and (2) *why* the robot took a particular decision. For each question, we defined two levels, as follows.

*What* was defined in terms of the information the robot needs to communicate to the user, designated here as 'verbosity.' In the low level of verbosity, short and possibly ambiguous explanations were provided, whereas the high level of verbosity provided precise and detailed explanations. *Why* was related to justification. In the low level of justification, no justification was provided. In the high level, precise justification for the robot's action was provided.

We thus developed four different LOEs using the two levels of verbosity and the two levels of justification, as shown in Table I.

TABLE I. DESIGN OF DIFFERENT LOEs

| LOE | Verbosity | Justification |
|---|---|---|
| Low | Low | Low |
| Medium1 | High | Low |
| Medium2 | Low | High |
| High | High | High |

## III. EXPERIMENTAL DESIGN

### A. The Robot and the Task

A Sawyer collaborative robot (cobot; 7 DOF) robotic arm was employed in a collaborative sorting task. The robot was required to sort the cubes onto one of two shelves according to the QR code it identified when picking the cube. The robot detected the cube's location, approached it, read the QR code, and sorted it accordingly. If the robot encountered an error, explanations were provided by the robot to the user. Upon generation of this dialog, the user was required to interact with the robot to rectify the error.

The interaction was conducted via an interface designed on a tablet mounted on top of the robot (Fig. 1). An RGB camera mounted on the end effector of the robotic arm was utilized to detect the QR code. Hand-eye calibration was implemented so that when the robot detected an object with a QR code it would pick it up and put it on the correct shelf. Two errors were deliberately introduced to create a need for user involvement in different conditions. Each participant experimented with the two errors presented at random. The first error was an incorrect item, i.e., an error occurred if the robot detected a QR code that was not in its database. The user was then expected to swap the cube with the correct one. The second error was an out-of-

Both authors are with the Department of Industrial Engineering and Management and the ABC Robotics Initiative, Ben-Gurion University of the Negev, Beer-Sheva 84105, Israel, shikhar@post.bgu.ac.il; yael@bgu.ac.il

*This work was partially supported by Ben-Gurion University of the Negev through the Agricultural, Biological, and Cognitive Robotics Initiative (funded by the Marcus Endowment Fund and the Helmsley Charitable Trust) and the Rabbi W. Gunther Plaut Chair in Manufacturing Engineering. We acknowledge the contributions of Gabreal Haj and Adam Eisa who developed the robot application. The robot was donated by HAHN robotics.

range item, which was defined as a cube placed out of the reach of the robot. The user was then expected to move the cube to a reachable location.

*B. The Conversational AI System*

We utilized the AI conversational system, Rasa[1], to develop an adaptive dialog system to generate explanations and thereby to improve the robot's understandability. Upon detection of an error by the robot, the Rasa module was automatically activated and the robot generated a low LOE. The low LOE was displayed on the tablet of the robot as well as on the GUI of the user's computer. The user was given two options on the GUI, i.e., either to press the continue button or ask any question of the robot. If the user asked, "*what is the error?*" or a similar question, then the robot would respond with a medium1 LOE (Table I). If the user asked, "*why has the error occurred?*" or a similar question, the robot would respond with either medium2 LOE or the high LOE.

If the user entered any other question or comment, the robot would respond by generating text such as: "I am sorry, please ask different question."

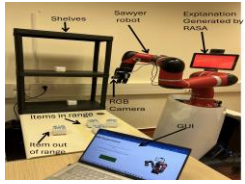

Fig. 1. Photograph of the robotic system connected to the GUI.

Two different adaptive dialog systems were designed. Each dialog started at low LOE (low verbosity and low justification). In the first adaptive dialog, denoted as adaptive dialog 1 (AD1) when the user questioned the '*what*,' the robot advanced to the medium1 LOE (high verbosity and low justification). Then, when the user questioned the '*why*,' the robot transitioned to medium2 LOE (low verbosity and high justification).

In the second adaptive dialog, adaptive dialog 2 (AD2) when the user questioned the '*what*,' the robot responded in medium1 LOE (high verbosity and low justification); when the user questioned the '*why*,' it responded in high LOE (high verbosity and high justification).

In both cases, the transition from one level to another was made through one parameter at a time (either verbosity or justification) and was not sequential.

## IV. RESULTS

Ten participants were recruited to demonstrate the feasibility of using the automatic dialog system and to compare the two dialog systems[2]. The results revealed that half the participants were able to resolve the error in adaptive dialog 1 and most participants (8/10) were able to resolve the error in adaptive dialog 2. A sample response is presented in Table II, and the remaining results are available at github[3].

[1] https://rasa.com/docs/
[2] Procedures were approved by the Department's ethics committee.
[3] https://github.com/shikharkumar1993/Adaptive-dialogue-sawyer-robot

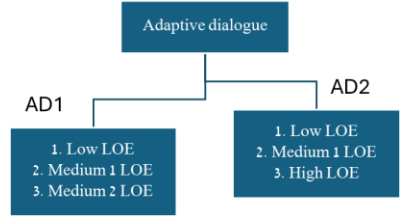

Fig. 2. The different adaptive dialog systems

TABLE II. SAMPLE OF DIALOG RESPONSE

| Error | User Queries | Response |
|---|---|---|
| Wrong cube | (Upon detection of error by the robot) | Error occurred (Low) |
| | What is the mistake with this | Error, I am unable to put the item on the shelf (Medium1) |
| | Why are you not able to place the cube | Error due to incorrect item. Swap the cube (Medium2) |
| Out-of-range | (Upon detection of an error by the robot) | Error occurred (Low) |
| | What is the error | Error I'm unable to reach the item on table (Medium1) |
| | Why are you not able to reach the cube | Error I'm unable to reach the item on the table because it is outside my camera vision. Please move it inside the square (High) |

## V. CONCLUSIONS AND FUTURE WORK

We designed four different LOEs (low, medium1, medium2 and high) based on verbosity (*what* the robot needs to explain) and justification (*why* the robot took a particular decision).

An adaptive dialog system was developed for transition from one LOE to another by utilizing a conversational AI agent. The adaptive dialog system was shown to enhance the understandability of the robot, enabling it to resolve errors with user assistance.

Ongoing work is aimed at extensive user studies comparing different types of adaptive dialogs using the above four LOEs for different collaborative tasks and conditions and evaluating additional LOEs for different questions (e.g., *when*, *how*).